\documentclass{article}
\usepackage{amsmath}
\usepackage{graphicx} 
\graphicspath{ {fig1/} }
\usepackage{booktabs} 
\usepackage{multirow}
\usepackage{cite}
\usepackage{geometry}
\usepackage{caption}

\title{\textbf{A dynamic vision sensor object recognition model based on trainable event-driven convolution and spiking attention mechanism}}

\newcommand{\super}[1]{\textsuperscript{#1}}
\author{%
  Peng Zheng\  \super{a,b,c}, \textrm{Qian Zhou} \super{a,b,c,*}%
}

\date{\textit{\super{a} State Key Laboratory of Reliability and Intelligence of Electrical Equipment,} \\
\textit{Hebei University of Technology, Tianjin, 300130, China} \\
\textit{\super{b} Tianjin Key Laboratory of Bioelectromagnetic Technology and Intelligent Health,} \\
\textit{Hebei University of Technology, Tianjin, 300130, China} \\
\textit{\super{c} Hebei Key Laboratory of Bioelectromagnetics and Neural Engineering,} \\
\textit{Hebei University of Technology, Tianjin, 300130, China}}

\usepackage{indentfirst}
\begin{document}
\maketitle

\begin{abstract}
Spiking Neural Networks (SNNs) are well-suited for processing event streams from Dynamic Visual Sensors (DVSs) due to their use of sparse spike-based coding and asynchronous event-driven computation. To extract features from DVS objects, SNNs commonly use event-driven convolution with fixed kernel parameters. These filters respond strongly to features in specific orientations while disregarding others, leading to incomplete feature extraction. To improve the current event-driven convolution feature extraction capability of SNNs, we propose a DVS object recognition model that utilizes a trainable event-driven convolution and a spiking attention mechanism. The trainable event-driven convolution is proposed in this paper to update its convolution kernel through gradient descent. This method can extract local features of the event stream more efficiently than traditional event-driven convolution. Furthermore, the spiking attention mechanism is used to extract global dependence features. The classification performances of our model are better than the baseline methods on two neuromorphic datasets including MNIST-DVS and the more complex CIFAR10-DVS. Moreover, our model showed good classification ability for short event streams. It was shown that our model can improve the performance of event-driven convolutional SNNs for DVS objects.
\end{abstract}

\begin{center}
\end{center}

\section{Introduction}

Inspired by the efficient information processing of the biological visual cortex using spikes, DVS output events independently and asynchronously at microsecond resolution for pixel locations where brightness changes exceed the threshold. Compared to traditional frame-based vision sensors, DVS has the characteristics of high temporal resolution, low latency, and high dynamic range. Because DVS are not affected by under/over-exposure or motion blur, they are more robust in low-light and highly dynamic scenes\cite{1}. In recent years, DVS has found wide application in target tracking and recognition\cite{2}, optical flow estimation\cite{3}, and other areas\cite{4,5}.

Although DVS has many potential advantages over frame-based visual sensors, most existing computer vision algorithms are designed for frame-based visual sensors. They use floating-point values and cannot directly process the sparse event stream data output by DVS. Therefore, most existing feature extraction methods for DVS objects involve overlaying (or transforming) events into a two-dimensional image representation as input to traditional neural networks for training and classification\cite{6,7}. These methods overlook the sparse nature of event streams and incur significant computational costs. Therefore, the effective extraction of feature information from sparse event streams is under-researched and poses a significant challenge to fully exploit their inherent advantages\cite{1}.

SNNs are the third generation of artificial neural networks which use sparse spikes to compute and propagate information. They mimic the spike-based and event-driven approach inherent in neural systems. By using sparse spikes to compute and propagate information, they have significant advantages in terms of low power consumption and hardware friendliness, thus emerges as an ideal neuromorphic computing paradigm\cite{8}. SNNs have great potential to provide efficient and low-latency solutions for event-based visual tasks\cite{9}. Compared to the conventional method, it is more natural for SNNs to process such sparse and temporal data by making full use of temporal features\cite{10}. These advantages have made SNNs natural candidates for processing the sparse, event-driven event stream data produced by DVS\cite{1}.

Event-driven convolution using oriented edge filters (e.g. Gabor, DOG filters) have been widely used in SNNs to extract DVS object features\cite{11}. The oriented edge filters emulate receptive fields of simple cells in the primary visual cortex, which respond strongly to features in a specific orientation. They can exploit the event-driven and low-latency advantages of DVS and improve the biointerpretability of the networks\cite{1,12}. However, their use fixed parameters means they respond strongly only to features of specific orientations while ignoring other features. This means that most SNNs using oriented edge filters are unable to fully extract event stream features, which limits their ability to achieve high classification performance for complex tasks\cite{13}.

Recently, some SNNs extracted global dependence features of event stream using attention mechanism\cite{14,15}. The attention mechanism is inspired by the ability of human visual cortex to dynamically restrict processing to a subset of the visual field\cite{16}. The spiking attention mechanism uses sparse spike computations for attention, as opposed to the traditional attention mechanism that relies on floating-point matrix multiplication. This approach offers clear advantages in terms of low computational cost and parameter efficiency\cite{17}. Like other attention-based Vision Transformer (VIT) models\cite{18}, this model focused exclusively on long-range dependencies between image patches. However, this can lead to a loss of information in spatial domain (such as edge features of objects in DVS) \cite{19,20}, which may limit the ability to extract features and perform well on complex tasks.

To overcome these limitations, this paper proposes a DVS object recognition model that consists of two parts: a trainable event-driven convolution module and a spiking attention module. The first part extracts local features from the event stream and updates the convolution kernel through gradient descent, enabling more efficient extraction of event stream features than traditional event-driven convolution. The second part extracts global dependence features from the input sequence and makes decisions by the firing rates of the output layer neurons. Our model has demonstrated significant improvements over baseline algorithms in the recognition of two popular AER datasets (MNIST-DVS and CIFAR10-DVS). Furthermore, our model has shown good classification ability for short event streams. In summary, the contributions of our work are as follows:
\begin{itemize}
\item We design a trainable event-driven convolution, where the convolutional kernel parameters are trainable during network training, rather than using fixed parameters of oriented edge filters. This advancement enhances the capability of event-driven convolution to extract features from the event stream.
\item After extracting DVS object features using the trainable event-driven convolution, we introduce a spiking attention mechanism to further extract global dependence features. Our approach achieves competitive results on two neural morphology datasets compared to the state-of-the-art methods.
\end{itemize}

The remaining sections of the paper are organized as follows: Section two introduces the existing work on object recognition with DVS; Section three describes the network architecture of our proposed model; Section four provides experimental details and results; Section five summarizes this work.

\section{Related Work}
\subsection{Event-driven convolution for DVS object recognition}
Inspired by the biological nervous system, SNN offer significant advantages in terms of low power consumption, making them naturally suitable for sparse object recognition with DVS. There are two main methods for extracting features and classifying objects with DVS include DNN-SNN-based and SNN-based. DNN-SNN-based methods require the conversion of asynchronous events into frame-based image representation as input to traditional neural networks for training and classification. Representative conversion methods include Event Frame\cite{21}, Voxel Grid\cite{22}, Event Spike Tensor\cite{23}. Traditional neural network training algorithms can be used with these methods to achieve high classification performance. However, they overlook the sparse and low-latency advantages of DVS\cite{24}. 

SNN-based methods are suitable for processing sparse, event-driven event stream data. However, traditional neural network training algorithms cannot be used due to the non-differentiability of the spike function. Some researchers proposed agent gradients\cite{25,26}, unsupervised STDP rules\cite{1,27}, and heterogeneous network\cite{28,29}to address this challenge. In addition, there are some studies on DVS object recognition using SNN that utilize the hierarchical processing of visual information in the ventral pathway (V1-V2-V4-IT) in primates. These studies used event-driven convolution in SNN to extract DVS object features and perform classification. Zhao\cite{29} proposed an event-driven SNN for DVS object recognition. The network directly extracts primary features of event stream data through event-driven convolution with Gabor filters. Feature classification is accomplished by using Tempotron learning rules. The achieved classification accuracies are 88.14\% and 99.48\% on the MNIST-DVS and AER Posture datasets, respectively. Xiao\cite{30} employed a similar SNN to Zhao\cite{29} and improved network classification accuracy by using multi-spike encoding in the feature encoding layer. The network achieved classification accuracies of 91.51\% and 99.95\% on the MNIST-DVS and AER Posture datasets, respectively. Liu\cite{31} used the feature extraction module of Zhao et al. 's SNN. They employed a SNN based on STDP rules for feature classification. Furthermore, the network introduced natural logarithmic encoding functions and multi-scale feature fusion in the feature encoding layer to improve classification accuracy. The network achieved accuracies of 89.96\%, 99.58\%, and 99\% on the MNIST-DVS, AER Posture, and Poker-DVS datasets, respectively. Orchard\cite{32} proposed a hierarchical SNN (HFirst) with a structure similar to the HMAX model for DVS object recognition. The network used a MAX operation based on the winner-take-all mechanism (WTA) in the C1 layer and employed statistical methods for feature classification. It achieved a classification accuracy of 97.5\% in a four-class card task. However, most of these studies are based on event-driven convolution using Gabor filters to extract oriented edge features from event streams. These methods elicit strong responses for specific directional features due to the fixed parameters of Gabor filters\cite{13}.

\subsection{Attention mechanism for DVS object recognition}
The attention mechanism is inspired by the dynamic ability of the human visual nervous system to focus on a specific region of the visual field\cite{15}. It has been widely applied in natural language processing, image processing, and other fields due to its emphasis on the relationship between local and global information\cite{33}. Researchers have explored the application of the attention mechanism for feature extraction and classification in DVS object recognition due to its outstanding performance in image recognition tasks.

Cannici\cite{34} used the attention mechanism to track event activity and localize regions of interest for DVS object recognition. However, this approach requires the reconstruction of sparse events into frame images, which overlooks the sparse and low-latency advantages of DVS. Peng\cite{35} considered the sparse event properties of DVS and proposed the Group Event Transformer model to use self-attention mechanism for DVS object recognition. The model achieved a classification performance of 99.7\% on the N-MNIST dataset. Cai\cite{36} proposed the Spatial-Channel-Temporal Fusion Attention (SCTFA) model, which effectively guides SNN to capture potential target regions by utilizing accumulated spatial-channel information. The model achieved a classification performance of 98.72\% on the MNIST-DVS dataset. To address the "data hunger" issue brought about by the attention mechanism, Li\cite{19} designed the Convolutional Transformer (CT) module. Furthermore, to better integrate the attention mechanism and the inherent spatio-temporal information of SNN, they adopted a spatio-temporal attention mechanism for DVS object recognition. The network achieved a classification performance of 90.97\% on the DVS-Gesture dataset. Zhou\cite{17} introduced a spiking-based attention mechanism that uses spiking-form Q, K, and V to calculate attention scores and extract global dependence features from input data. They model was applied to the recognition of static and neural morphology datasets. However, these methods rely on attention mechanisms to capture the correlation of event stream data from DVS. The attention mechanism enables the model to pay more attention to the most informative components of the input, potentially overlooking spatial domain information such as edge features of DVS objects\cite{19,20}.

\section{Method}
In this section, we introduce the proposed DVS object recognition model. The framework of the model is shown in Fig. 1. The model consists of two parts: a trainable event-driven convolution module and a spiking attention module.
\subsection{Trainable event-driven convolution module}
We propose the trainable event-driven convolution module, which counts the number of times each parameter of the event-driven convolution kernel is covered at each position on the response map event-by-event. The module extracts the event stream features using the trained convolution kernel.

Traditional event-driven convolution\cite{11} is a commonly used method for feature extraction of DVS objects. It has the advantage of high temporal resolution and low latency. Its convolution kernel is centered on the event position, and the response map is updated by adding to the current response map. As shown in Fig. 2a and 2c, assuming the index of the convolution kernel, the response map, the convolution kernel, and the events in the response map. The traditional event-driven convolution process is shown in Fig. 2b. However, due to the fixed-parameter convolution kernel, it only responds strongly to specific edge features and cannot fully extract event stream features. We found that the values of each position of the response map can be expressed as a linear superposition of each parameter in the convolutional kernel after processing event-driven 
\begin{figure}[!t]
  \centering
  \includegraphics[width=16cm, height=6cm]{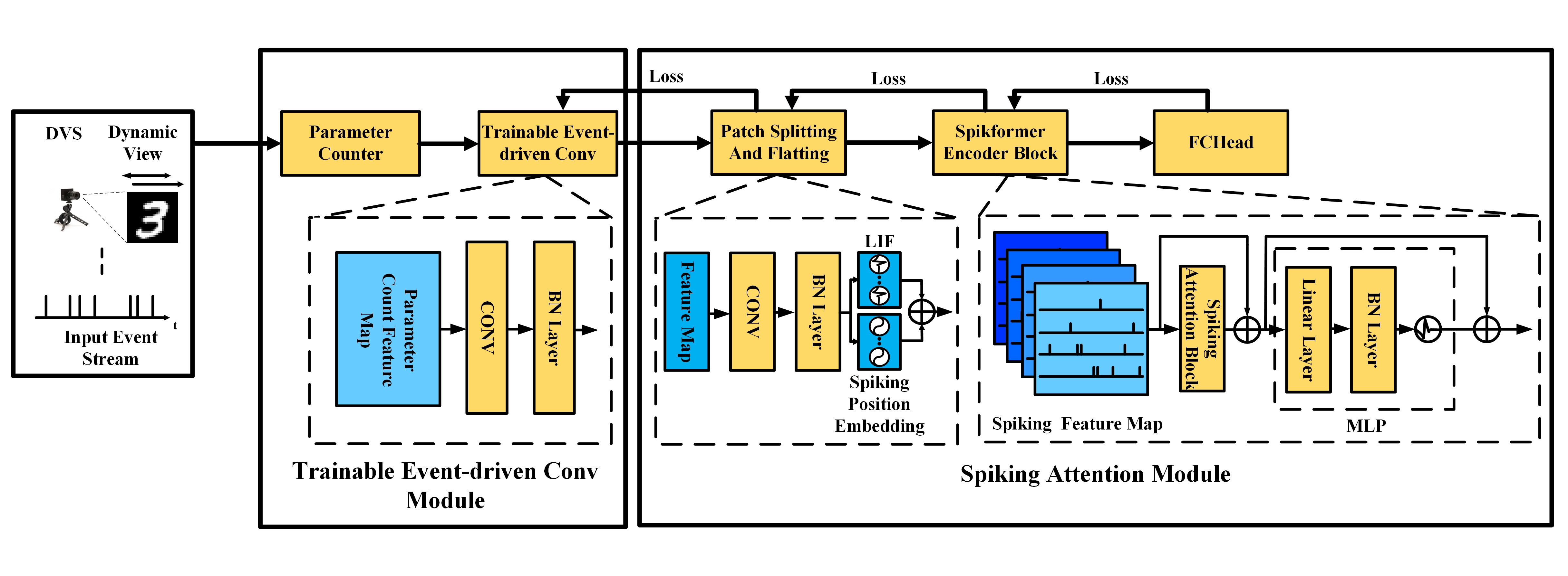}
  \caption{The overall structure of the proposed model based on trainable event-driven convolution and spiking attention mechanism}
  \label{fig:标签名称}
\end{figure}
convolution for an event stream. For instance, the value at position A of the response map in Fig. 2b can be expressed as Eq. (1):
\begin{equation}
Responemap({\rm{A}}) = {n_{AW1}} \times {W_1} + {n_{AW2}} \times {W_2}_{} +  \cdots  + {n_{AWn}} \times {W_n}
\end{equation}
where ${W_n}$ denotes the parameter with position index $n$ on the convolution kernel, and ${n_{AWn}}$ denotes the number of times that A position has been covered by the convolution kernel parameter ${W_n}$ after processing the raw event stream.

It indicates that if we can obtain the number of times each convolutional kernel parameter covers each position on the response map and perform one convolution with a stride equal to the event-driven convolution kernel size, we will obtain the response map of the event-driven convolution. This is equivalent to the response map obtained by performing convolution event by event, as shown in Fig. 2b and 2d. Therefore, the convolutional kernel can be updated through gradient descent during the model training process. During the inference phase, the model can extract DVS object features using the trained convolution kernel. This method can extract features of the event stream more effectively than traditional event-driven convolution.

The trainable event-driven convolution proposed in this paper consists of a parameter counter and a trainable convolutional layer. The parameter counter counts the number of times each parameter in the convolutional kernel covers each position on the response map event by event. As shown in Fig. 2d, when the parameter counter receives the event ${e_1} = ({t_1},{x_1},{y_1})$, the parameter counter centers on the position $({x_1},{y_1})$ and the parameter count feature map is updated by adding 1 to the value of the corresponding position. It means that AW1, BW2, CW3, FW4, GW5, HW6, KW7, LW8, and MW9 are updated by adding 1. Similarly, for the event ${e_2} = ({t_2},{x_2},{y_2})$, the parameter counter centers on the position $({x_2},{y_2})$, RW1, SW2, TW3, WW4, XW5, and YW6 are updated by adding 1. The final map obtained from the parameter counter in this way is called the parameter count feature map, which is fed to the trainable convolutional layer. One convolution is performed on the parameter count feature map, the convolution stride is the size of the event-driven convolution kernel. Note that this layer is involved in the model training process, and the convolution kernel is updated during training to achieve smaller loss function values.
\begin{figure}[!t]
  \centering
  \includegraphics[width=17.2cm, height=9cm]{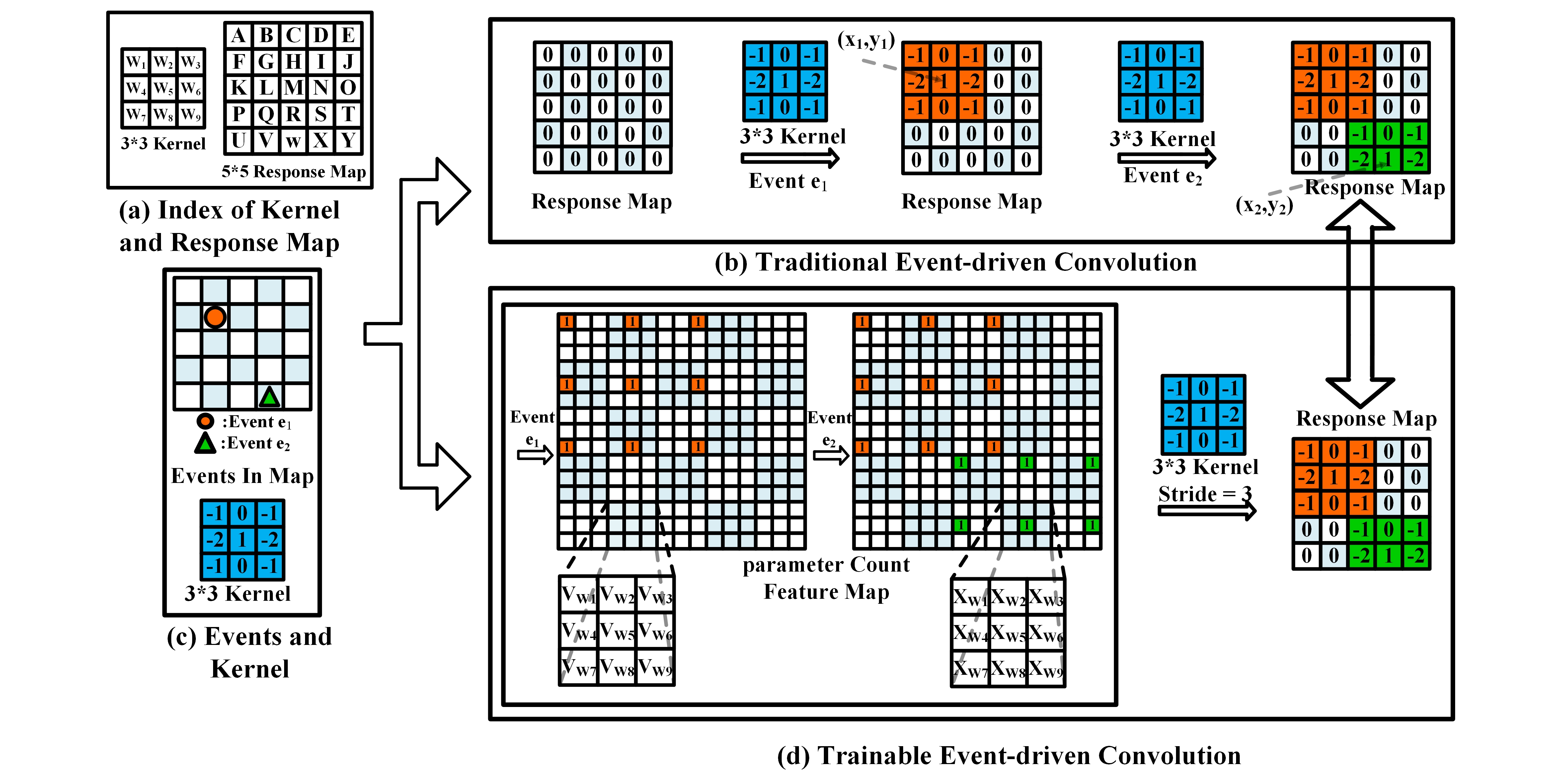}
  \caption{Traditional event-driven convolution and trainable event-driven convolution. (a) The index of the convolution kernel and the response map. (b) Traditional event-driven convolution. (c) The convolution kernel and the events in the response map. (d) Trainable event-driven convolution}
  \label{fig:标签名称}
\end{figure}
\subsection{Spiking attention module}
After extracting the features using the trainable event-driven convolution module, we use spiking attention mechanism to extract the global dependence features of the event stream. The spiking attention module consists of a patch split and flattened block, a spikformer encoder block, and a fully connected classification head.
\subsubsection{Patch split and flattened block}
By introducing patches into the attention mechanism, models can effectively focus on relevant parts of the input image, leading to more efficient and effective processing\cite{36}. Since the spiking attention mechanism only receives patches\cite{17}, we use patch split and flattened block to convert the response map of the trainable event-driven convolution module to patches. This process involves a convolutional layer and a LIF neuron layer. The convolutional layer maps the response map to the patches, while the LIF neuron layer encodes the patches as spiking sequences $x$. We use the conditional position encoder\cite{37} to characterize the relative position information of the spiking sequences $x$. The encoder encodes the spiking sequence as the relative position spiking sequence ${x_{pe}}$, which is then directly summed with the spiking sequence $x$ to obtain the spiking sequence ${x_o}$. The process of patch splitting and flattening is as Eq. (2):

\begin{equation}
x = \text{LIF}(\text{Conv2d}(Featuremap)) \quad x \in \text{Spike}_s^{T \times N \times D}
\end{equation}
\begin{equation}
x_{pe} = \text{LIF}(\text{Position embedding}(x)) \quad x_{pe} \in \text{Spike}_s^{T \times N \times D}
\end{equation}
\begin{equation}
x_o = x + x_{pe} \quad x_o \in \text{Spike}_s^{T \times N \times D}
\end{equation}

\subsubsection{Spikformer encoder block}
The spikformer encoder block consists of a spiking attention block and a variable number of multilayer perceptrons (MLPs) to extract global dependence features. The spiking attention block receives ${x_o}$ from the patch split and flattened block. Float-point components $Q$, $K$, $V$ are computed by three sets of trainable weights ${W_Q}$, ${W_K}$, ${W_V}$. The LIF neurons encode the three key components as the spiking sequence ${Q_{Spike}}$, ${K_{Spike}}$, ${V_{Spike}}$. The attention map generated by the spiking attention block is inputted into the MLP, which is utilized to extract high-level features for classification purposes.
\subsubsection{Fully connected classification head}
The fully connected classification head receives the feature vector from the MLP to classify. It consists of a fully connected layer and a LIF neuron layer, the LIF neuron layer receives outputs from the fully connected layer and generates currents from dynamic equations. It accumulates membrane potentials and emits spikes when the membrane potential exceeds the threshold. The decision is according to the average frequency of spikes emitted by each neuron within a period. The label represented by the neuron is the final decision of the model.
\section{Experiment and Result Analysis}
\subsection{Experimental setup}
Our model was implemented using SpikingJelly\cite{17}, an open-source and high-speed simulation framework based on PyTorch. The model can efficiently run on GPU platforms. Our experiments were conducted on a server equipped with an NVIDIA GeForce RTX 3090TI GPU (24GB GPU memory) and an Intel Core i7-8700 3.20GHz CPU.

The performance of the proposed model was evaluated on two neuromorphic datasets: MNIST-DVS\cite{38}and CIFAR10-DVS\cite{39}. The MNIST-DVS dataset contains recordings from event cameras that capture the motion processes of 10 different classes of handwritten digit images (0-9) in the MNIST dataset. 10,000 digit images were randomly selected from a set of 70,000 frame-based $28 \times 28$ MNIST images. Each image was then individually scaled up using a smooth interpolation algorithm to three different scales. Next, a DVS was fixed in front of an LCD, and each handwritten digit image was made to move slowly across the display. Finally, the fixed event camera was used to record these slow-moving handwritten digits for about 2-3 seconds each. In our experiments, we used 10,000 event stream samples, with 1,000 samples for each digit type.

The CIFAR10-DVS dataset is a high-quality and challenging event-stream dataset that consists of 10 categories: aircraft, cars, birds, cats, deer, dogs, frogs, horses, boats, and trucks. The directional gradient of the frame-based image is converted to a local relative intensity change in the event camera's field of view using image repetitive closed-loop smoothing motion. This conversion was achieved by an event-based image sensor from the CIFAR-10 dataset. Each category contains 6,000 samples, resulting in a total of 60,000 event streams. The spatial resolution is $128 \times 128$, which is currently considered high resolution in common neuromorphic datasets\cite{40}. This dataset presents a greater challenge for the DVS object recognition model due to the larger spatial resolution and more complex object classes\cite{41}.

For each dataset, 90\% of the samples are randomly selected for training, and the remaining samples are used for testing. We have selected a subset of algorithms that have demonstrated state-of-the-art performance on these two datasets for comparison purposes. The chosen baseline methods include traditional deep learning methods and spiking neural networks that have achieved advanced or leading classification performance. These algorithms include a spiking neural network constructed using an improved spiking neuron model\cite{42,43}, the graph-based spiking neural network\cite{44,45}, the network that incorporates attention mechanisms\cite{15,17,36}, the spiking neural network trained with improved algorithms\cite{46,47,48,49,50}, and other approaches\cite{51,52,53,54,55}.
\subsection{Performance on neuromorphic datasets}
We evaluate the performance of the proposed model using the two selected neuromorphic datasets. For the MNIST-DVS dataset with input size of $28 \times 28$,  the trainable event-driven convolution module uses convolution kernel of size $3 \times 3$ and stride size 3. We use four spikformer encoder blocks., 6 attention heads, and 4 MLP blocks. The time steps are 12, the batch size is 10. For the CIFAR10-DVS dataset with input size of $128 \times 128$, we also use the convolution kernel of size $3 \times 3$ and stride size 3. There are two spikformer encoder blocks, 16 attention heads, and 4 MLP blocks. The time steps are set to 10, and the batch size is 8. Table 1 shows the classification performance of our model with existing state-of-the-art (SOTA) or comparable methods.

Our model achieves classification performance almost equivalent to the state-of-the-art method (LIAF-Net \cite{42}) for the MNIST-DVS dataset, with a difference of only 0.2\%. However, it uses 20 time steps, resulting in lower latency compared to another method, VA-Net\cite{15}, which uses 64 time steps. Our model uses only 12 time steps and still achieves comparable classification performance. The CIFAR10-DVS dataset is more challenging due to high intra-class differences. The model achieves classification performance beyond that of all baseline methods using short time steps (T=10). It improves performance by 5.4\% over the Dspike model \cite{48}, 3.5\% over the DSR model \cite{51}, and 1.9\% over the Spikformer model \cite{17} at the same time steps. The classification accuracies on both datasets demonstrate that the model has excellent classification performance with low latency.
\begin{table}[!t]
\captionsetup{justification=centering} 
\caption{Classification performances of our model and existing SOTA or comparable methods on MNIST-DVS and CIFAR10-DVS.}
\centering
\begin{tabular}{cccc}
\toprule 
Dataset & Methods & Time Steps & Acc \\
\midrule
\textbf{ }  & DART (TPAMI 2019) \cite{52}  & -   & 98.5\% \\
\textbf{ }    & RG-CNNs (ICCV 2019) \cite{45}   & -  & 98.6\%  \\
\textbf{ }    & EvS-B (ICCV 2021) \cite{44} \textbf{(SOTA)}   & -  & 99.1\%  \\
\textbf{ }    & SSFE (2024) \cite{54} \textbf{(SOTA)}   & -  & 99.1\%  \\
MNIST-DVS    & GEM-SNN (2022) \cite{53}   & 10  & 97.2\%  \\
\textbf{ }    & SCTFA (2023) \cite{36}   & 20  & 98.7\%  \\
\textbf{ }    & VA-Net (ICASSP 2022) \cite{15}   & 64  & 99.0\%  \\
\textbf{ }    & LIAF-Net (2021) \cite{42} \textbf{(SOTA)}   & 20  & 99.1\%  \\
\textbf{ }    & \textbf{Our}   & 10  & 98.9\%  \\
\midrule
\textbf{ }  & SALT (Neural Netw-2021) \cite{46}  & 20   & 67.1\% \\
\textbf{ }    & SFA (2023) \cite{49}   & 50  & 71.9\%  \\
\textbf{ }    & SEW-ResNet (NeurIPS-2021) \cite{47}   & 16  & 74.4\%  \\
\textbf{ }    & PLIF (ICCV-2021) \cite{43}   & 20  & 74.8\%  \\
CIFAR10-DVS    & Dspike (NeurIPS-2021) \cite{48}   & 10  & 75.4\%  \\
\textbf{ }    & DSR (CVPR-2022) \cite{51}   & 10  & 77.3\%  \\
\textbf{ }    & SLTT (ICCV-2023) \cite{50}   & 10  & 77.3\%  \\
\textbf{ }    & MPBN (ICCV-2023) \cite{55}   & 10  & 78.7\%  \\
\textbf{ }    & Spikformer (2022) \cite{17}   & 10  & 78.9\%  \\
\textbf{ }    & \textbf{Our}   & 10  & 80.8\%  \\
\bottomrule 
\end{tabular}
\end{table}
\subsection{Ablation studies}
\subsubsection{Trainable event-driven convolution module}
To verify the necessity of the trainable event-driven convolution module, we first compare the classification performances of the trainable event-driven convolution with the traditional event-driven convolution on the MNIST-DVS. For both methods, we use a full-length (approximately 2s) MNIST-DVS event stream data as input. The Gabor filter scale and the convolution kernel scale in the trainable event-driven convolution were both set to 3. The Gabor filter convolution kernel generation formula followed the method used by Serre \cite{56}, with an effective width $\sigma$ of 1.2, a wavelength $\lambda $ of 1.5, an aspect ratio $\gamma $  of 0.3, and orientations $\theta $ of 0°, 45°, 90°, and 135°. The fully connected layer comprised 784 neurons to receive the feature vectors extracted by the event-driven convolution. The network decision layer consisted of 10 neurons to make classification decisions.

Table 2 shows the classification performances of the traditional Gabor filter-based event-driven convolution combined with the fully connected layer and the trainable event-driven convolution combined with the fully connected layer.

\begin{table}[!h]
\centering
\captionsetup{justification=centering} 
\caption{Classification performances of traditional event-driven convolution by Gabor filter and \\trainable event-driven convolution on MNIST-DVS} 
\label{my-table} 
\begin{tabular}{cc} 
\toprule 
Methods & Acc \\
\midrule
Gabor + Fully connected   & 87.1\% \\
Trainable event-driven convolution + Fully connected  & 95.7\% \\
\bottomrule
\end{tabular}
\end{table}

The traditional event-driven convolution achieves an accuracy of 87.1\%, while trainable event-driven convolution achieves an accuracy of 95.7\%. The trainable event-driven convolution improves the performance of the traditional event-driven convolution by 8.6\%. This indicates that the trainable event-driven convolution module can effectively improve the feature extraction performance of the model.

We also compare the classification performance of our model with the model without the trainable event-driven convolution module. We use the same parameters in Section 4.2 for the classification model of the MNIST-DVS dataset. The classification performances are shown in Table 3.

\begin{table}[!h]
\centering
\captionsetup{justification=centering} 
\caption{Classification performances of our model and the model without \\the trainable event-driven convolution module} 
\label{my-table} 
\begin{tabular}{cc} 
\toprule 
Methods & Acc \\
\midrule
Spiking attention mechanism   & 95.8\% \\
Trainable event-driven convolution + Spiking attention mechanism  & 98.9\% \\
\bottomrule 
\end{tabular}
\end{table}

As shown in Table 3, the model without the trainable event-driven convolution module achieves an accuracy of 95.8\%, while our model achieves an accuracy of 98.9\%. Our model improves the performance of the model without the trainable event-driven convolution module by 3.1\% on MNIST-DVS. It shows that the trainable event-driven convolution module proposed in this paper can improve the feature extraction performance of the model.
\subsubsection{Spiking attention module}
To verify the necessity of the spiking attention module, we compare the classification performance of our model with the model without the spiking attention module. Since classification cannot be accomplished using the trainable event-driven convolution module alone, we use a fully connected layer to make network decisions. The model parameters are the same as the Section 4.2 for the MNIST-DVS dataset classification model. The classification performances are shown in Table 4.

\begin{table}[!h]
\centering
\captionsetup{justification=centering} 
\caption{Classification performances of our model and the model without \\the spiking attention module on MNIST-DVS} 
\label{my-table} 
\begin{tabular}{cc} 
\toprule 
Methods & Acc \\
\midrule
Trainable event-driven convolution + Fully connected   & 95.7\% \\
Trainable event-driven convolution + Spiking attention mechanism  & 98.9\% \\
\bottomrule 
\end{tabular}
\end{table}

As shown in Table 4, the model without the spiking attention module achieves an accuracy of 95.7\%, while our model achieves an accuracy of 98.9\%. Our model improves the performance of the model without the spiking attention module by 3.2\% on MNIST-DVS. This indicates that the spiking attention module can improve the feature extraction performance of the model. 
\subsection{Influence of the input event stream time length on model performance}
In this section, we analyze the influence of the event stream time length on the classification performance of our model. In our experiments, we employed complete event stream samples or their segments from two neural morphology datasets as input to the model. For the MNIST-DVS dataset, 100ms, 200ms, 500ms, and full length event stream fragments are used as input. We use the same parameters in Section 4.2 for the MNIST-DVS dataset classification model. For the CIFAR10-DVS dataset, 200ms, 500ms, and full length event stream fragments are used as input. We also use the same parameters in Section 4.2 for the CIFAR10-DVS dataset classification model.

The classification performances of our model and the model without the trainable event-driven convolution module with different time lengths are shown in Table 5.
\begin{table}[!htbp]
\centering
\captionsetup{justification=centering} 
\caption{Classification performances of our model and model without the trainable event-driven convolution module on MNIST-DVS and CIFAR10-DVS with different time lengths}
\label{tab:my_label}
\begin{tabular}{cccc}
\toprule
Datasets & Event Stream Time Length (ms) & Spiking Attention Mechanism & Our  \\
\midrule
\multirow{4}{*}{MNIST-DVS} & 100 & 92.5\% & 97.2\% \\
 & 200 & 94.0\% & 97.9\% \\
 & 500 & 95.4\% & 98.8\% \\
 & full & 95.8\% & 98.9\% \\
\midrule
\multirow{3}{*}{CIFAR10-DVS} & 200 & 72.7\% & 76.8\% \\
 & 500 & 76.8\% & 79.1\% \\
 & full & 78.9\% & 80.4\% \\
\bottomrule
\end{tabular}
\end{table}

As shown in Table 5, the performance of our method is always the best regardless of the duration of the event stream. When using the MNIST-DVS dataset segments of 100ms, our model significantly outperforms the model using the spiking attention module alone with 97.2\% accuracy. For the complex CIFAR10-DVS dataset, our model achieves superior classification accuracy even when using only 500ms segments, and outperforms the model using only the spiking attention mechanism module on full-length segments. This indicates that our model has excellent feature extraction and recognition performance when using short event streams.
\subsection{Influence of the network architecture on model performance}
This section analyses the influence of event stream time length on our model's classification performance. The model is fed complete event stream samples or their segments from two neural morphology datasets in our experiments. For the MNIST-DVS dataset, we use event stream fragments of 100ms, 200ms, 500ms, and full length as input. The same parameters as in Section 4.2 are used for the MNIST-DVS dataset classification model. The CIFAR10-DVS dataset employs 200ms, 500ms, and full-length event stream fragments as input. The same parameters as in Section 4.2 are used for the CIFAR10-DVS dataset classification model.

The classification performances of our model with different network architectures are shown in Table 6.
\begin{table}[!htbp]
\captionsetup{justification=centering} 
\caption{Classification performances of our model on MNIST-DVS and CIFAR10-DVS with different architectures} 
\centering
\begin{tabular}{ccc} 
\toprule 
Datasets & Number of Spikformers & Acc \\
\midrule
\multirow{3}{*}{MNIST-DVS} 
& 1 & 97.8\% \\
& 2 & 98.1\% \\
& 4 & 98.9\% \\
\midrule
\multirow{2}{*}{CIFAR10-DVS} 
& 1 & 80.4\% \\
& 2 & 80.8\% \\
\bottomrule 
\end{tabular}
\end{table}

Table 6 shows that our model's classification performance for the MNIST-DVS dataset improves by 1.1\% when using 4 encoding blocks compared to 1 encoding block. Similarly, for the CIFAR10-DVS dataset, the classification performance improves by 0.4\% when using 2 encoding blocks compared to 1 encoding block. Overall, increasing the number of encoding blocks leads to a corresponding improvement in classification performance.
\section{Conclusion}
This paper presents a DVS object recognition model that overcomes the limitations of inadequate feature extraction caused by traditional event-driven convolution. The model uses a trainable event-driven convolution module to extract features of the DVS object and a spiking attention module to extract global dependence features. Competitive performances on both neuromorphic datasets have been achieved by our model. The results of our extensive experiments demonstrate the necessity of both parts in our model. Furthermore, our model exhibits good classification ability for short event streams, which demonstrates its superior ability to extract the features of DVS objects.
\section*{CRediT authorship contribution statement}
\textbf{Peng Zheng}: Investigation, Conceptualization, Methodology, Validation, Software, Formal analysis, Visualization, Writing original draft. \textbf{Qian Zhou}: Resources, Writing- review \& editing, Supervision, Data curation.
\section*{Declaration of Competing Interest}
The authors declare that they have no known competing financial interests or personal relationships that could have appeared to influence the work reported in this paper.
\section*{Acknowledgments}
This work is supported by the National Natural Science Foundation of China (Grant No. 61305077).

\bibliographystyle{IEEEtran}
\bibliographystyle{unsrt} 


\end{document}